\documentclass[10pt,twocolumn,letterpaper]{article}

\usepackage{cvpr}
\usepackage{times}
\usepackage{epsfig}
\usepackage{graphicx}
\usepackage{amsmath}
\usepackage{amssymb}
\usepackage{subfigure}
\usepackage{bm}
\usepackage{mathrsfs}
\usepackage{url}
\usepackage{algorithm}
\usepackage{algorithmic}

\usepackage[breaklinks=true,bookmarks=false]{hyperref}

\cvprfinalcopy 

\def\cvprPaperID{****} 

\begin{document}

\title{Multi-Modality Fusion based on Consensus-Voting and 3D Convolution for Isolated Gesture Recognition}

\author{Jiali Duan$^1$, Shuai Zhou\thanks{Joint first author} , Jun Wan\thanks{Corresponding author} , Xiaoyuan Guo$^{2}$, and Stan Z. Li$^{3}$\\
\small $^{1,\dag,3}$Center for Biometrics and Security Research \& National Laboratory of Pattern Recognition \\
\small Institute of Automation, Chinese Academy of Sciences \\
\small $^*$ Macau University of Science and Technology \\
\small $^{2}$ School of Engineering Science, University of Chinese Academy of Sciences\\
{\tt\small $^{1,*,2}$\{jli.duan,shuaizhou.palm,xiaoyuanguo.ucas\}@gmail.com, $^{\dag,3}$\{jun.wan,szli\}@nlpr.ia.ac.cn}}



\maketitle

\begin{abstract}
Recently, the popularity of depth-sensors such as Kinect has made depth videos easily available while its advantages have not been fully exploited. This paper investigates, for gesture recognition, to explore the spatial and temporal information complementarily embedded in RGB and depth sequences. We propose a convolutional two-stream consensus voting network (2SCVN) which explicitly models both the short-term and long-term structure of the RGB sequences. To alleviate distractions from background, a 3d depth-saliency ConvNet stream (3DDSN) is aggregated in parallel to identify subtle motion characteristics. These two components in an unified framework significantly improve the recognition accuracy. On the challenging Chalearn IsoGD benchmark, our proposed method outperforms the first place on the leader-board by a large margin (10.29\%) while also achieving the best result on RGBD-HuDaAct dataset (96.74\%). Both quantitative experiments and qualitative analysis shows the effectiveness of our proposed framework and codes will be released to facilitate future research.

\end{abstract}

\section{Introduction}

Vision based gesutre recognition has drawn much attention from both the academic and industrial community for its widespread applications, such as human computer interaction and sign language translation and so on. Many methods have been proposed over the last few years~\cite{beyond-gesture, online-gesture, 3d-gesture,jun-CVPRW}, which can be classified mainly into two categories: continuous gesture recognition and isolated gesture recognition. The former can be converted to the latter once continous gestures are segmented into seperate ones.

\begin{figure}[htb]
\centering
\subfigure{
\label{rgb1}
\includegraphics[width=0.09\textwidth]{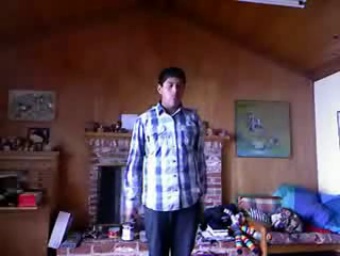}}
\hspace{-1ex}
\subfigure{
\label{rgb2}
\includegraphics[width=0.09\textwidth]{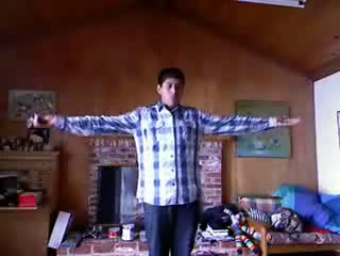}}
\hspace{-1ex}
\subfigure{
\label{rgb3}
\includegraphics[width=0.09\textwidth]{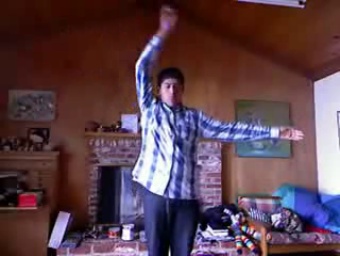}}
\hspace{-1ex}
\subfigure{
\label{rgb4}
\includegraphics[width=0.09\textwidth]{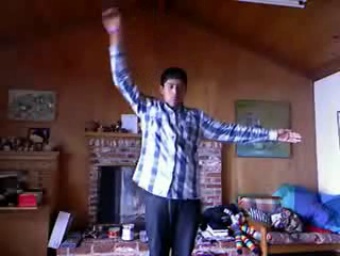}}
\hspace{-1ex}
\subfigure{
\label{rgb4}
\includegraphics[width=0.09\textwidth]{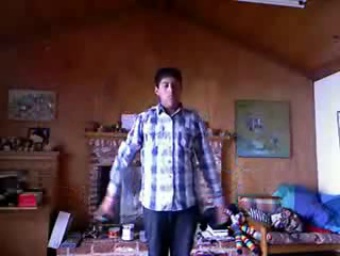}} \\
\vspace{-1ex}
\subfigure{
\label{depth1}
\includegraphics[width=0.09\textwidth]{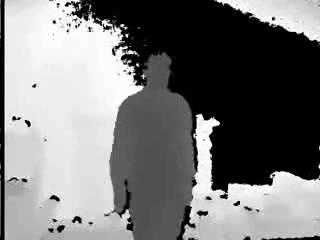}}
\hspace{-1ex}
\subfigure{
\label{depth2}
\includegraphics[width=0.09\textwidth]{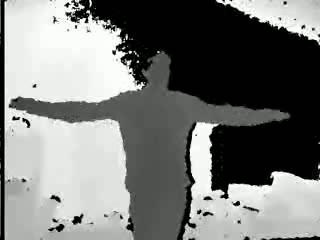}}
\hspace{-1ex}
\subfigure{
\label{depth3}
\includegraphics[width=0.09\textwidth]{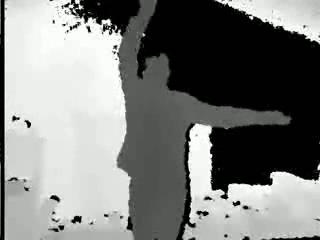}}
\hspace{-1ex}
\subfigure{
\label{depth4}
\includegraphics[width=0.09\textwidth]{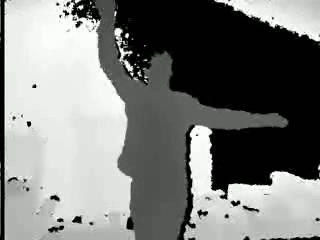}}
\hspace{-1ex}
\subfigure{
\label{depth4}
\includegraphics[width=0.09\textwidth]{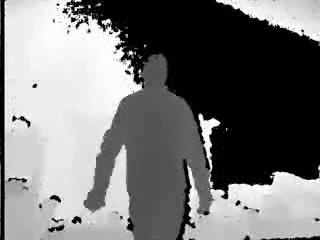}} \\
\vspace{-1ex}
\subfigure{
\label{flow1}
\includegraphics[width=0.09\textwidth]{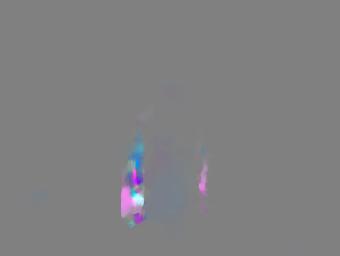}}
\hspace{-1ex}
\subfigure{
\label{flow2}
\includegraphics[width=0.09\textwidth]{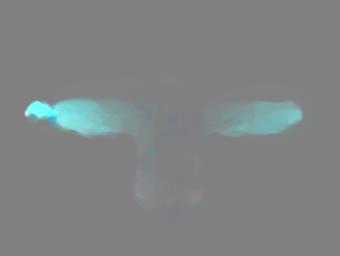}}
\hspace{-1ex}
\subfigure{
\label{flow3}
\includegraphics[width=0.09\textwidth]{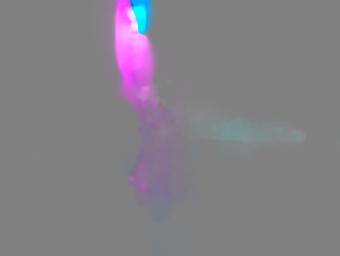}}
\hspace{-1ex}
\subfigure{
\label{flow4}
\includegraphics[width=0.09\textwidth]{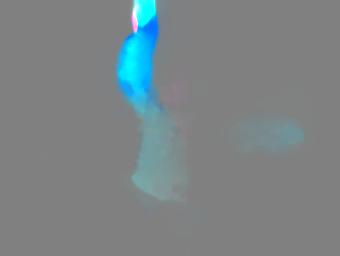}}
\hspace{-1ex}
\subfigure{
\label{flow5}
\includegraphics[width=0.09\textwidth]{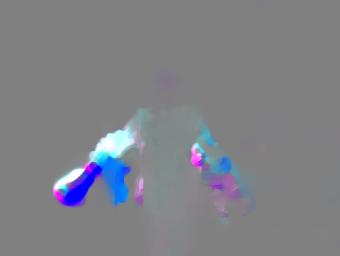}} \\
\vspace{-1ex}
\subfigure{
\label{sal1}
\includegraphics[width=0.09\textwidth]{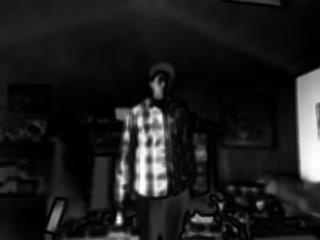}}
\hspace{-1ex}
\subfigure{
\label{sal2}
\includegraphics[width=0.09\textwidth]{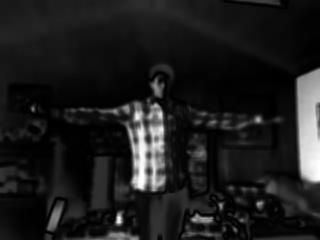}}
\hspace{-1ex}
\subfigure{
\label{sal3}
\includegraphics[width=0.09\textwidth]{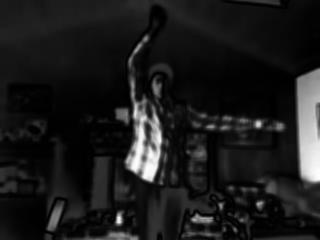}}
\hspace{-1ex}
\subfigure{
\label{sal4}
\includegraphics[width=0.09\textwidth]{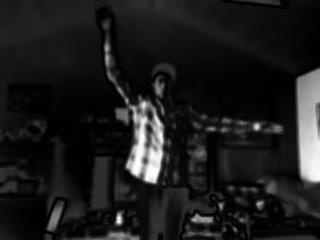}}
\hspace{-1ex}
\subfigure{
\label{sal4}
\includegraphics[width=0.09\textwidth]{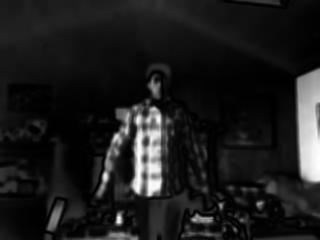}}
\caption{Examples of different types of modalities listed from up to bottom: RGB images, Depth images, Optical flow fields (magnitudes obtained from x and y direction are used as color channel), and Saliency}
\label{modality}
\end{figure}

In this paper, we focus on isolated gesture recognition, especially for RGB-D video sequence, an area that has recently come into popularity due to the advancement and availability of depth-sensors such as Kinect. Although a significant amount of efforts have been made in the area of video recognition~\cite{beyond-short-snippet,vector-cnn,improve-trajectory} for RGB modality, the complementary advantages inherently embedded in RGB-D sequences have largely been ignored. As shown in Fig.\ref{modality}, depth sequence contains structural information from the depth channel and are more capable of dealing with noises from background, clothing, skin color and other external factors, thereby concentrating on the salient regions i.e. gestures. As were pointed out in recent works such as~\cite{depth2action,NTU-RGBD,wan2016explore}, depth cues could act as important supplement to the original RGB sequence, particularly for datasets that contain subtle inter-class variations. Besides depth information, we also include saliency for inspection.
 \begin{figure*}[thb]
  \centering
  \scalebox{1}{
  \includegraphics[width=0.90\textwidth]{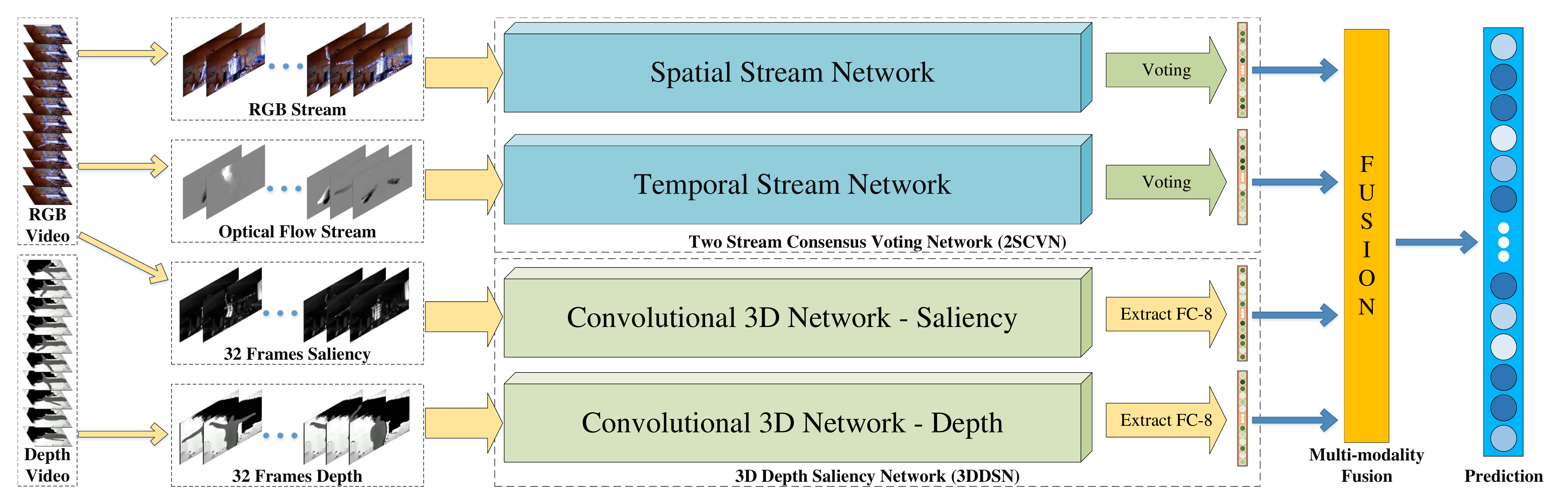}}
  \caption{An overview of our approach. An input video is represented in different modalites, where the RGB stream and optical flow fields are handled by \emph{Two Stream Consensus Voting Network (2SCVN)} while the saliency-stream and depth-stream are handled by \emph{3D Depth-Saliency Network (3DDSN)}. 2SCVN takes into account consensus votes of the first two streams and outputs scores, which are later fused with another two streams from 3DDSN to give the final scores.}
  \label{fig:overview}
\end{figure*}
In fact, an impressive performance gain brought about by these two modalities as explained in Section~\ref{sec:exploration} further consolidates our observation.

Our main motivations are: \emph {1) how to reduce estimation variance when it comes to the decision of classifying videos of comprehensive inter-and-intra class vairations; 2) how to design a general and effective framework that is able to take advantage of different modalities.}

For the first problem, we notice that unlike other video recognition tasks such as action recognition, which contains relatively rich contextual information of body correlations and interactions, the task of gesture recognition usually involves only the motion of hands and arms. In other words, existing gesture recognition methods~\cite{online-gesture,3d-gesture,beyond-gesture} which deal with a limited number of gestures can make very ``biased'' estimations when it comes to classifiying gesture datasets that involve comprehensive inter-and-intra class variations. Second, current main-stream approaches such as~\cite{two-stream, LRCN} usually deal with short-term motions, possibly missing important information from actions that span over a relatively long time. For example, some gestures such as ``OK'' or number signals involve only motions of a short period while gesticulations denoting forced landing, diving signals or slow motions require temporal modeling of a relatively long sequence.

To solve the aforementioned issue, we propose a novel two stream network (2SCVN) based on the idea of consensus voting adapted from~\cite{two-stream,TSN,two-stream-fusion}. It first takes frames sampled from different segments of the sequence according to uniform distribution and stacks their corresponding optical flow fields as input. Compared to dense sampling or pre-defined sampling interval which may be highly redundant,  this leads to less computations and ensures that videos which are short or those which involve multiple stages can be completely covered fairly well. These frames are then combined to cover more diversity before being fed into the spatial and temporal streams of 2SCVN for video level predictions. Finally, these predictions are aggregated to reduce estimation variance.

For the second problem, we realize that as human motions are in essense three-dimensional, the information loss in the depth channel could cause degradations to the discriminative capability of feature representation. On the other hand, saliency helps eliminate ambiguity from possible distractions of color camera. To the best of our knowledge, we are the first to perform investigations that highlight spatial and temporal combinations from these two modalities, based on which 3D depth-saliency (3DDSN) fusion scheme is proposed. Eventually, predictions from both 2SCVN and 3DDSN are taken into consideration as the final score. What's worth noticing is that our proposed approach also works surprisingly well for other tasks of video recognition (See Table \ref{tab:all-result}), demonstrating the effectiveness and generalization ability of our framework.

The proposed framework is shown in Fig.\ref{fig:overview} and the main contributions of our paper are:

\begin{figure*}[htb]
  \centering
  \includegraphics[width=0.99\textwidth]{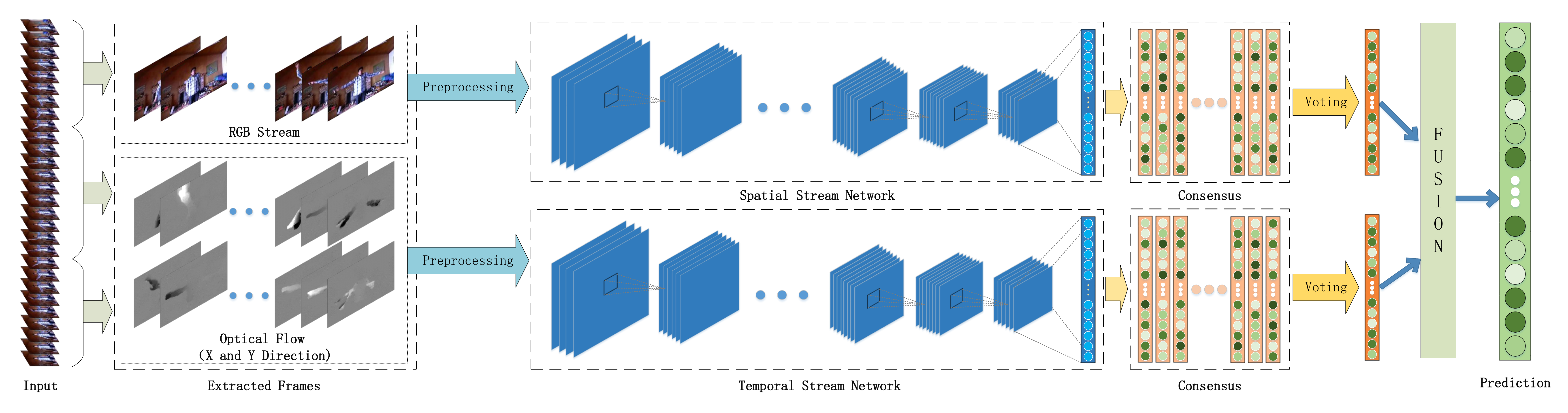}
  \caption{2SCVN is based on the idea of consensus voting, where its spatial and temporal stream sample RGB images and its stacked optical flow fields from different segments of a sequence according to a uniform distribution. Consensus from these frames as well as their augmented counterparts are taken as ``votes'' for the predictions.}
  \label{fig:2SCVN}
\end{figure*}

\begin{enumerate}
\item We proposed a novel framework that combines the merits of 2SCVN and 3DDSN for multi-modality fusion. It absorbs depth and saliency streams as important constituents to capture subtle spatial-temporal information supplementary to RGB sequence.
\item A convolutional network design (2SCVN) based on the idea of consensus voting is proposed to explicitly model the long term structure of the whole sequence, where video-level predictions from each frame and its augmented counterparts are aggregated to reduce possible estimation variance.
\item We are the first to perform an integration of 3D depth-saliency stream to address the loss of three-dimensional structural information and distractions from backgrounds, noises and other external factors.
\item Our approach performs particularly well not only for RGB-D gesture recognition but also for human daily action recognition, achieving the best results on ChaLearn IsoGD~\cite{jun-CVPRW} and RGBD-HuDaAct~\cite{hudaact} benchmarks.
\end{enumerate}

The remainder of this paper is organized as follows: Section~\ref{sec:related_work} is an review of related works. Our unified framework is illustrated in Section~\ref{sec:approach} and validated in Section~\ref{sec:exploration} respectively. Section~\ref{sec:conclusion} concludes the paper.


\section{Related Work}
\label{sec:related_work}
In this section, we first introduce some works for gesture recognition deploying hand-crafted features, then we review works related to ours both in the field of gesture and action recognition that conduct research on different modalities and convolutional networks.

Many hand-crafted features have been proposed for video analysis in the area of gesture recognition~\cite{Starner1998Real, Wang2006Hidden, Dardas2011Real,Hand-ges1,Wan2013One}. For example, Wan et al~\cite{wan2016explore} extracted a novel spatiotemporal feature named MFSK while~\cite{Wan20143D} proposed to calculate SIFT-like descriptors on 3D gradient and motion spaces respectively for RGB-D video recognition. Dardas et al~\cite{Dardas2011Real} recognized hand gestures via bag-of-words vector mapped from extracted keypoints using SIFT and a multiclass SVM was trained as gesture classifier.

Recently, the convolutinal neural networks~\cite{LeCun-cnn} have been introduced to the field of gesture and action recognition due to its rich capacity for representation~\cite{3d-gesture,Nishida2015Multimodal,online-gesture}. Additionally, the rapid emergence of depth-sensor has made it economically feasible to capture both color and depth videos, providing motion information as well as three-dimensional structual information. This significantly reduces motion ambiguity when projecting the three-dimensional motion onto the two-dimensional image plane~\cite{hudaact,NTU-RGBD,Wan2014CSMMI}. For example, Molchanov et al~\cite{3d-gesture} proposes to use depth and intensity data with 3D convolutional networks for gesture recognition. Nishida et al~\cite{Nishida2015Multimodal} proposes a multi-stream recurrent neural network that can be trained end to end without domain-specific hand engineering while~\cite{online-gesture} combines 3DCNN with RNN for online gesture detection and classification. Ohn-Bar et al~\cite{Ohn2014Hand} first detects a hand in the region of interaction and then combines RGB and depth descriptor for classification. Neverova et al~\cite{Neverova2015Multi} proposes a multi-modal architecture that operates at 3 temporal scales corresponding to dynamic poses for gesture localization.

In the camp of action recognition, Karpathy et al~\cite{large-video-classification} extended CNNs into video classification on a large-scale dataset of 1 million videos (Sports-1M).
Donahue et al.~\cite{LRCN} embraced recurrent nerual networks to explicitly model the complex temporal dynamics. Tran et al.~\cite{3d-convolution} proposed to simultaneously extract the spatio-temporal features with deep 3D Convolutional Neural Networks (3D-CNN) followed by a SVM classifier for classification.
Simonyan et al.~\cite{two-stream} designed an architecture that captures the complementary information on appearance and motion between frames. Based on which, Feichtenhofer et al.~\cite{two-stream-fusion} studied several levels of granularity in feature abstraction to fuse spatial and temporal cues.


\section{Our Method}
\label{sec:approach}


Fig.\ref{fig:overview} is an overview of our proposed approach. It mainly consists of Two Stream Consensus Voting Network (2SCVN) and 3D Depth-Saliency Network (3DDSN). Votings from 2SCVN and Fc-8 outputs from 3DDSN represent predictions from different modalities. These scores are further fused as the eventual label for isolated gesture recognition.

\subsection{Two Stream Consensus Voting Network}

As is pointed out in Introduction, the bottleneck for improving the performance of large-scale gesture recognition lies in: 1) comprehensive inter-and-intra class variations; 2) long-term modeling of motions from sequences of variable length. Here we base our method on top of mainstream approaches~\cite{two-stream,two-stream-fusion} and adopts an \emph{Consensus Voting Strategy} to reduce estimation variance.

\textbf{Consensus Voting Strategy:}
\label{sec:consensus}
The structure of 2SCVN is illustrated in Fig.\ref{fig:2SCVN}. We formalize the operations by convolutional networks as $F$ parameterized by $\theta $:

\begin{equation}
F:{\Re ^{h \times w \times t \times m}} \to {\Re ^l},\bm{f} = F(\tau ;\theta )
\end{equation}

\noindent where an input snippet $\tau$ of sequential length $m \ge 1$ with t channels of size $h \times w$ pixels is transformed into a vector $\bm{f}$. Then, we apply softmax function $g:{\Re ^l} \to {\Re ^l}$ on top of vector $\bm{f}$

\begin{equation}
{[g(\bm{f})]_i} = {e^{{\bm{f}_i}}}/\sum\nolimits_k {{e^{{\bm{f}_k}}}}
\end{equation}

\noindent where the $i^{th}$ dimension indicates the probability of the snippet belonging to class $i$. Therefore, given an input video $V$ of $T$ snippets, we can calculate ${[p({c_1}|{\tau _j}),p({c_2}|{\tau _j})...p({c_l}|{\tau _j})]^T}$, the probability with respect to each category for snippet ${\tau _j}$. By stacking these predictions together, we get the following matrix:

\[\left[ {\begin{array}{*{20}{c}}
{p({c_1}|{\tau _1})}&{p({c_1}|{\tau _2})}&{ \cdot  \cdot  \cdot }&{p({c_1}|{\tau _T})}\\
{p({c_2}|{\tau _1})}&{p({c_2}|{\tau _2})}&{ \cdot  \cdot  \cdot }&{p({c_2}|{\tau _T})}\\
{ \cdot  \cdot  \cdot }&{ \cdot  \cdot  \cdot }&{ \cdot  \cdot  \cdot }&{ \cdot  \cdot  \cdot }\\
{p({c_l}|{\tau _1})}&{p({c_l}|{\tau _2})}&{ \cdot  \cdot  \cdot }&{p({c_l}|{\tau _T})}
\end{array}} \right] \xrightarrow{h} \left[ {\begin{array}{*{20}{c}}
{p({c_1}|V)}\\
{p({c_2}|V)}\\
{ \cdot  \cdot  \cdot }\\
{p({c_l}|V)}
\end{array}} \right]\]

\noindent where each column is the class predictions of each snippet and each row being the class-specific predictions from $T$ snippets. The aggregation function (\emph{voting}) $h :{\Re ^{l \times T}} \to {\Re ^l}$ then combines the predictions from each snippet along the horizontal axis to output the probability of the whole video $V$ with respect to each class. Therefore, the predicted label for video $V$ is

\begin{equation}
y = \mathop {\arg \max }\limits_{i \in {S_l}} (p(c|V))
\end{equation}

Note that the choice of $h$ is still an open question and is determined by each specific task, here we have tried out Max and Mean funciton in Section~\ref{sec:aggregate}.

Using the prediction of video $V$ for each class, we deploy the standard categorical cross-entropy loss to train our network:

\begin{equation}\label{2SCVN:loss}
L(\bm{y},\bm{p}) =  - \sum\limits_{i = 1}^l {{y_i}({p_i} - \log \sum\limits_{j = 1}^l {{e^{{p_j}}}} )}
\end{equation}

\noindent where $l$ is the number of categories and ${y_i}$ the ground truth label concerning class $i$.
Each network parameter with respect to the loss fuction is updated by stochastic gradient descent with a momentum $\mu  = 0.9$. Each parameter in the network $\theta  \in \omega$ is updated at every iteration step $t$ by

\begin{equation}
{\theta _t} = {\theta _{t - 1}} + {\nu _t} - \gamma \lambda {\theta _{t - 1}}
\end{equation}

\begin{equation}
{\nu _t} = \mu {\nu _{t - 1}} - \lambda \eta \left( {{{\left\langle {\frac{{\delta L}}{{\delta \theta }}} \right\rangle }_{batch}}} \right)
\end{equation}

where $\lambda $ is the learning rate, $\gamma $ is the weight decay parameter and $ < \delta L/\delta \theta { > _{batch}}$ is the gradient of cost function $L$ with respect to parameter $\theta$ averaged over the mini-batch. To prevent gradient explosion, we apply a soft gradient clipping operation $\eta $~\cite{gradient-clip}.

\textbf{Implementations:} We conducted experiments on Inception~\cite{Inception} with respect to the choice of ConvNet architecture due to its good balance between efficiency and accuracy. However, training deep networks is challenged by the risk of over-fitting as current datasets for video recognition are relatively small compared to other computer vision tasks such as image classification. A common practice is to initialize the weights with pre-trained models on ImageNet~\cite{Imagenet}. To further mitigate the problem, we also adopted batch-normalization~\cite{Batch-norm} and dropout~\cite{dropout} layer for regularization. Data augmentation is also employed to cover the diversity and variability of training samples. Besides random cropping and horizontal flipping, we also adapted the scale-jittering cropping technique~\cite{VGG} to involve not only jittering of scales but also aspect ratios.

The optical flow fields are acquired using~\cite{Wedel2009An}. We use Caffe~\cite{jia2014caffe} to train our networks. The learning rate is set to 0.1 and decreases to its $1/10$ for every 1500 iterations, lasting for over 20 epochs. It takes about 6 hours and 22 hours for training the spatial and temporal stream respectively on ChaLearn IsoGD with 2 TITANX GPUs.

\subsection{3D Depth-Saliency Network}
\label{sec:3DDSN}

\begin{figure}[htb]
  \hspace{-1.5ex}
  \includegraphics[width=0.49\textwidth]{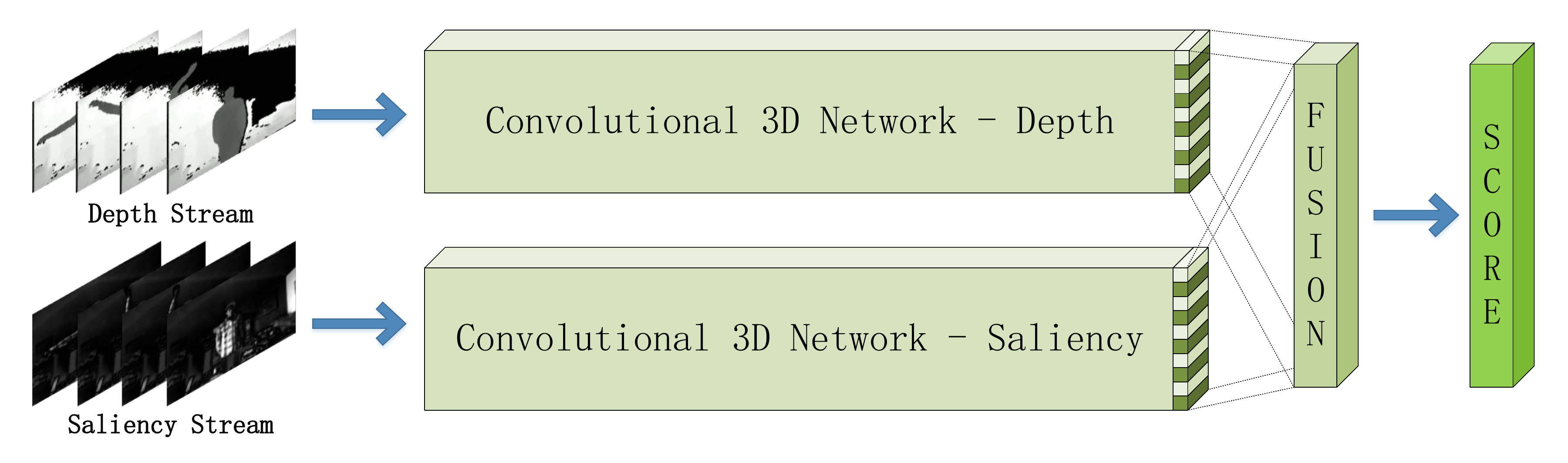}
  \caption{3DDSN employs 3D convolution on depth and saliency stream respectively, then takes scores from each stream for late fusion.}
  \label{fig:3DDSN}
\end{figure}

\textbf{Network Architecture:} We base our method on top of 3D convolutional kernel proposed by Tran et al~\cite{3d-convolution} while getting rid of the orginal Linear SVM configuration to train in an end to end manner. Compared to previous deep architectures, 3D CNNs are capable of encoding the spaital and temporal information in the data without requiring additional temporal modeling. Fig.~\ref{fig:3DDSN} shows the structure of 3DDSN.

Specifically, we propose to combine depth and saliency stream, based on the observation that depth incorporates 3-dimensional structural information that RGB doesn't while saliency helps reduce the influence from backgrounds and other noises so as to focus on the salient regions, see Fig.\ref{modality} for illustration. Each stream consists of eight 3D convolutional layers, each with a nonlinear Relu layer followed by five 3D Max Pooling layer. More formally, the 3D convolutional layer of the spatial-temporal CNN is defined as

\begin{equation}\label{3D:Convolution}
\sum\limits_{{\delta _t}} {\sum\limits_{{\delta _y}} {\sum\limits_{{\delta _x}} {{F_{t + {\delta _t}}}(x + {\delta _x},y + {\delta _y})} } }  \times \omega ({\delta _x},{\delta _y},{\delta _t})
\end{equation}

where $x$ and $y$ define the pixel position for a given frame ${F_t}$. Then, nonlinearities are injected with Rectified linear unit, followd by the 3D pooling layer, defined as follows

\begin{equation}\label{3D:Relu}
{\mathop{\rm Re}\nolimits} Lu(x,y,t) = \left \{ {\begin{array}{*{20}{l}}
{\begin{array}{*{20}{l}}
{Conv(x,y,t)} &{ Conv > 0}
\end{array}}\\
{\begin{array}{*{20}{l}}
0&{} &{} &{} &{} &{} &{otherwise}
\end{array}}
\end{array}} \right.
\end{equation}

\begin{equation}\label{3D:Pooling}
Pool(x,y,t) = \mathop {\max }\limits_{x,y,t} ({\mathop{\rm Re}\nolimits} Lu(x,y,t))
\end{equation}

Let $\chi  = \{ {V_0},{V_1},...,{V_B}\}$ be a mini-batch of training samples and $\omega$ be the network's parameters. During training, we append a softmax prediction layer to the last fully-connected layer and finetune back-propagation with negative log-likelihood to predict classes from individual video $V_i$

\begin{equation}\label{3d-loss}
L(\omega ,\chi ) =  - \frac{1}{{|\chi |}}\sum\limits_{i = 0}^{|\chi |} {\log (p({c^{(i)}}|{V^{(i)}};\omega ))}
\end{equation}

where ${p({c^{(i)}}|{V^{(i)}})}$ is the probability of class label ${{c^{(i)}}}$ given video ${{V^{(i)}}}$ as predicted by 3DCNN. Finally, the predictions from the depth and saliency stream are fused to give the eventual label.

\textbf{Implementations:} Saliency images are extracted using~\cite{3d-saliency}. We first re-sampled each sequence to 32 frames using nearest neighbour interpolation by droping or repeating frames~\cite{resample}. Given each frame $F_t$, volumes are constructed using its surrouding 32 frames (${F_{t - 15:t + 16}}$) with the label being the gesture occuring at its central frame. The spatial-temporal kernel size is set to $3 \times 3 \times 3$ in our experiments and the sacle of the pooling is set to $2 \times 2 \times 2$ for all but the first layer. Addionally, the generalization ability of deep learning methods relies heavily on the data it trains on. In the specific task of gesture recognition, we observe that users might randomly choose their left or right hands while performing a gesture without changing the meaning, thus we adopt horizontal flipping as agumentation technique to incorporate this varaiability. The network is finetuned on sports-1M model with base learning rate of 0.0001 (decrease to its $1/10$ every 5000 stepsize) for 100K iterations. It needs about 2 days to finetune and update parameters and takes about 8G graphic memory for each modality.

\section{Experiments}
\label{sec:exploration}
To tap the full potential of our unified framework for RGB-D gesture recognition, we have explored extensively with various settings to examine how each component influences the final performance and experimented a number of good practices in terms of data augmentation, regularization and model fusion. We also visualize the confusion matrix, to give an intuitive analysis.

\subsection{Datasets}

RGB-D gesture recognition datasets suitable for evaluation of deep-learning based methods are very rare. Therefore, besides evaluations on ChaLearn IsoGD gesture recognition dataset\cite{jun-CVPRW}, we also conducted experiments on RGBD-HuDaAct~\cite{hudaact}, one of the largest RGB-D action recognition datasets, where our proposed approach beat other methods, achieving state-of-the-art results.

\textbf{Chalearn IsoGD:} The CharLearn LAP RGB-D Isolated Gesture Dataset (IsoGD) contains 47933 RGB-D two-modality video sequences manually labeled into 249 categories, of which 35878 samples belong to the training set. Each RGB-D video represents one gesture instance, having 249 gesture labels performed by 21 different individuals. The IsoGD benchmark is one of the latest and largest RGB-D gesture recognition benchmarks and has a clear evaluation protocol, upon which the 2016 ChaLearn LAP Large-scale Isolated Gesture Recognition Challenge has been held~\cite{chalearn}. For the following evaluations, we conduct our experiments and report our accuracies on this dataset if not specifically mentioned.


\textbf{RGBD-HuDaAct:} The RGBD-HuDaAct database aims to encourage research efforts on human activity recognition on multi-modality sensor combination and each video is synchronized with color and depth streams. It contains 1189 samples of 13 activity classes (including background videos which are added to the exsiting 12 classes) performed by 30 volunteers with rich intra-class variations for each activity representation.

In the following subsections, we follow the pipeline of Algorithm~\ref{alg:pipeline} for testing of our proposed approach.

\begin{algorithm}[!htb]\small   
\caption{Test pipeline of our proposed approach}
\label{alg:pipeline}
\begin{algorithmic}[1]
\REQUIRE~\\
\textbf{Input and Model:} 2SCVN and 3DDSN model; \\
RGB-D video I \\
\ENSURE \textbf{label}
\STATE $I   \xrightarrow{divide}  S=\{ {S_1},{S_2},{S_3},...{S_k}\} $
\STATE $V=\{\}$
\FOR {$\ {S_t} \ in \ S  $}
\STATE ${\tau _t} \xleftarrow{sample} S_t$
\STATE  $V= V \cup {\tau _t}$
\ENDFOR
\STATE $V \xleftarrow{augment} V$
\STATE $RGB, Flow, Depth, Sal \xleftarrow{compute} V $
\STATE $RGB \xrightarrow{2SCVN} Spatial\ Votes $
\STATE $Flow \xrightarrow{2SCVN} Temporal\ Votes $
\STATE $p\{2SCVN\} \xleftarrow{h} \{Sptial\ Votes \cup Temporal \ Votes\}  $
\STATE $Depth \xrightarrow{3DDSN} Depth \  Scores $
\STATE $Saliency \xrightarrow{3DDSN} Depth \  Scores $
\STATE $p\{3DDSN\} \xleftarrow{h} \{Depth \  Scores \cup Depth \  Scores\}  $
\STATE $p \xleftarrow{multi-modality} \{p\{2SCVN\}\cup p\{3DDSN\}\}$
\STATE $y = \mathop {\arg \max }\limits_{i \in {S_l}} (p(c|V))$
\RETURN label
\end{algorithmic}
\end{algorithm}

\subsection{Aggregation Function Discussion}
\label{sec:aggregate}

In this subsection, we focus on discussions related to 2SCVN. As is mentioned in Section~\ref{sec:consensus}, aggregation function used for ``voting'' $h$ is an open problem and is determined by a specific task. Here we empirically evaluted two kinds of functions, max and mean. Table~\ref{tab:2SCVN} shows the accuracies of the spatial and temporal stream of 2SCVN under different aggregation functions.

\begin{table}[htb]\small
\centering
 \setlength{\abovecaptionskip}{-2pt}
 \setlength{\belowcaptionskip}{-2pt}
 \caption{Accuracies of different modalities and their combinations in the framework of 2SCVN on ChaLearn IsoGD, where Max and Mean aggregation functions have been tested. ``-F'' indicates optical flow fields}
\label{tab:2SCVN}
  \begin{center}
  \begin{tabular}{l || c | c | c }
    \hline
    2SCVN      & RGB   &  RGB-F             &RGB+RGB-F           \\
    \hline
    Max        & 45.65\%    & 58.36\%       & 62.72\%                     \\
    Mean       & 43.52\%   & 56.74\%        & 61.23\%                    \\
    \hline
   \end{tabular}
   \end{center}
\end{table}

From Table~\ref{tab:2SCVN}, we can draw the following conclusions: 
1) The performance of temporal stream is higher compared to spatial stream, which is reasonable because the spatial stream only captures actions at a fixed frame while the temporal stream takes into consideration motions at different time steps; 2) In terms of ``voting'', max aggregation seems to be more effective than mean aggregation and we leave other aggregation approaches as future work for related fields; 3) Compared to each modality, the combination of spatial and temporal stream leads to improvement in performance.

\subsection{Fusion Schemes}

In this subsection, we focus on discussions related to 3DDSN and explored the following questions: 1) How do depth, saliency perform individually; 2) Whether feature fusion does better than score fusion; 3) Any need for pre-processing before fusion? We conduct experiments on the Chalearn IsoGD benchmark using the aforementioned network configuration for each stream and explored their combinations.

\begin{table}[htb]\small
\hspace{-2ex}
 \setlength{\abovecaptionskip}{0pt}
 \setlength{\belowcaptionskip}{-4pt}
 \caption{Accuracies of different modalities on ChaLearn IsoGD, ``-'' indicates that softmax is not used while``+'' indicates vice-versa. D is short for Depth while S for saliency. }
\label{tab:3d}
  \begin{center}
  \begin{tabular}{l || c | c | c }
    \hline
    3DCNN        & Depth             &Saliency    & D:Sal(2/1)       \\
    \hline
    Softmax -    & 54.95\% & 43.36\%              &  58.86\%                            \\
    Softmax +    & 54.95\% & 43.36\%              &  56.37\%                                \\
    \hline
   \end{tabular}
   \end{center}
\end{table}

We trained the mainstream 3D + SVM approach~\cite{3d-convolution} as our baseline and used the same network architecture mentioned above, except that the spatial-temporal features from depth and saliency streams were concatenated to train the SVM classifier. The training of SVM takes about 6 hours and the accuracy on IsoGD~\cite{jun-CVPRW} is 53.60\%. Table~\ref{tab:3d} reports the highest accuracies on IsoGD benchmark of different modalities and their combinations according to the evaluation protocol. The following conclusions can be derived: 1) Depth seems to be the more effective and discriminative than saliency; 2) Although for one modality, whether or not using softmax to convert the output to range $[0-1]$  yields the same accuracy, it generally reports higher accuracies for modality fusion without the "softmax" pre-processing. This is perhaps that the conversion reduces variance of features, thus abating the discriminative ability of model ensemble; 3) Compared with 3D + SVM baseline which employes feature-level fusion, score fusion seems to be more preferable, since features from different modalities may have very different distributions, therefore simple concatenation is not valid.

\subsection{How does depth matter?}
Besides recognition accuracies, to get a full appreciation of the potential from depth information, we compared RGB and RGB + Depth model trained using the architecture mentioned in Section~\ref{sec:3DDSN} and  counted the changes after fusing the depth into rgb stream and depth bring changes to ``Correct'' and ``Error''.

\begin{figure}[htb]\small
  \hspace{-2ex}
  \includegraphics[width=0.50\textwidth]{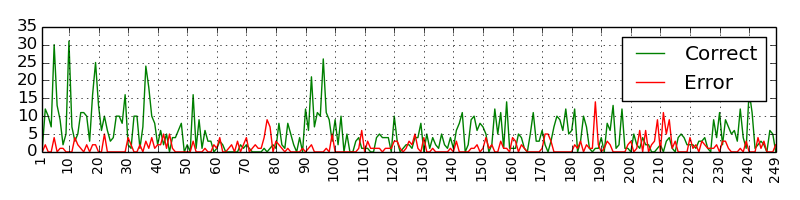}
  \caption{Changes after fusing the depth stream into RGB stream. The x-axis denotes the category ID while the y-axis represents the number of changes. For an individual ID, ``Correct'' means that rgb stream makes the wrong predictions but rgb+depth fusion are correct. Conversely, ``Error'' indicates that the number of changes when the vice-versa is true.}
  \label{fig:changes}
\end{figure}

\begin{figure*}[htb]
  \centering
  \subfigure[RGB]{
  \label{2scvn:rgb}
  \includegraphics[width=0.18\textwidth]{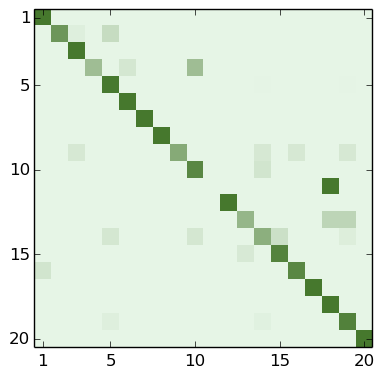}}
  \subfigure[RGB-Flow]{
  \label{2scvn:flow}
  \includegraphics[width=0.18\textwidth]{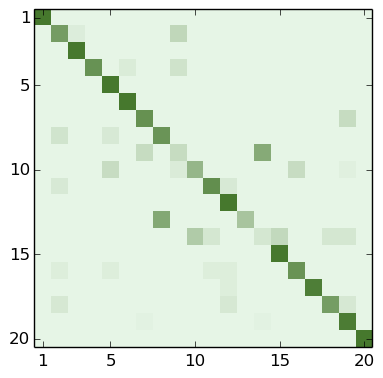}}
  \subfigure[Depth]{
  \label{3d:depth}
  \includegraphics[width=0.18\textwidth]{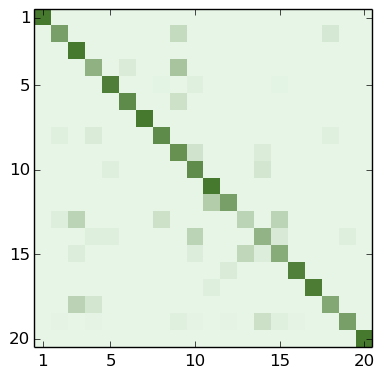}}
  \subfigure[Saliency]{
  \label{3d:sal}
  \includegraphics[width=0.18\textwidth]{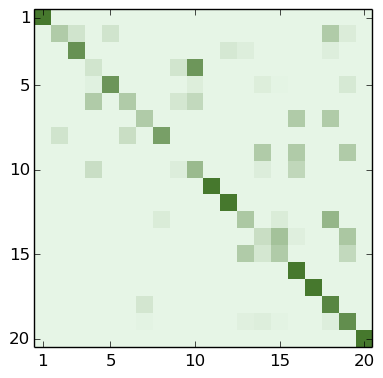}}
  \subfigure[proposed]{
  \label{all:fusion}
  \includegraphics[width=0.18\textwidth]{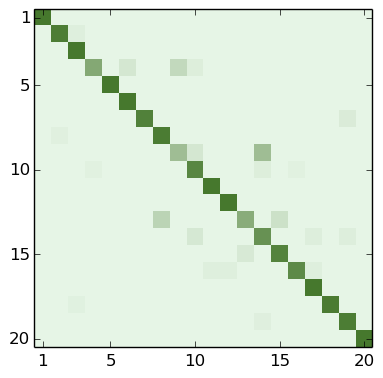}}
  \hspace{1ex}
  \subfigure{
  \label{3d:colormap}
  \includegraphics[height=0.18\textwidth]{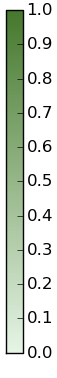}}
  \caption{Performance confusion matrix of 2SCVN for RGB and RGB-F and 3DDSN for Depth and Saliency as well as the proposed fusion model on ChaLearn IsoGD dataset. The first 20 categories are used for visulization due to page size and the whole confusion matrix can be inferred from supplementary materials.}
  \label{fig:conf-compare}
\end{figure*}

A big ``Correct/Error'' means that depth brings positive/negative effect on RGB stream while zero means depth has no effect on the final prediction of that class. As shown in Fig.\ref{fig:changes}, RGB stream works well in the range of $110-140$, however there are some class ranges such as $90$ to $100$, we have seen a huge improvement in terms of correct changes brought about by depth stream. The higher the green line (``Correct''), the more samples which have originally been predicted wrong are now correct. As the height of the green line is generally higher than that of red line, it confirms that depth indeed provides important supplementary information to RGB stream. Note that as the RGB stream of 3DDSN performs worse than that of 2SCVN, therefore this stream is not adopted in our final framework.

\subsection{Visualization of Confusion Matrix}

Fig.\ref{fig:conf-compare} displays the confusion matrix of RGB and RGB-Flow, Depth, Saliency and overall approach.

From Classes such as $9$ in RGB stream (Fig.\ref{2scvn:rgb}) are not misclassified while there exist some confusions in RGB-Flow (Fig.\ref{2scvn:flow}). On the other hand, classes such as $11$ which are confused in RGB stream perform relatively well in RGB-Flow. Thus, the spatial and temporal information actually supplements each other.


The confusion matrix of \emph{Depth} and \emph{Saliency} stream from 3DDSN on ChaLearn IsoGD are shown in Fig.\ref{3d:depth} and Fig.\ref{3d:sal} respectively. Fig.\ref{all:fusion} shows the confusion matrix of our proposed approach after modality fusion, which is obviously better than seperate streams. Note that we only displayed the first 20 categories due to page size and the whole confusion matrix are available in supplementary materials.

\subsection{Qualitative Results}

\begin{figure*}[htb]
  \hspace{4ex}
  \subfigure{
    \label{re1}
    \includegraphics[width=0.22\textwidth]{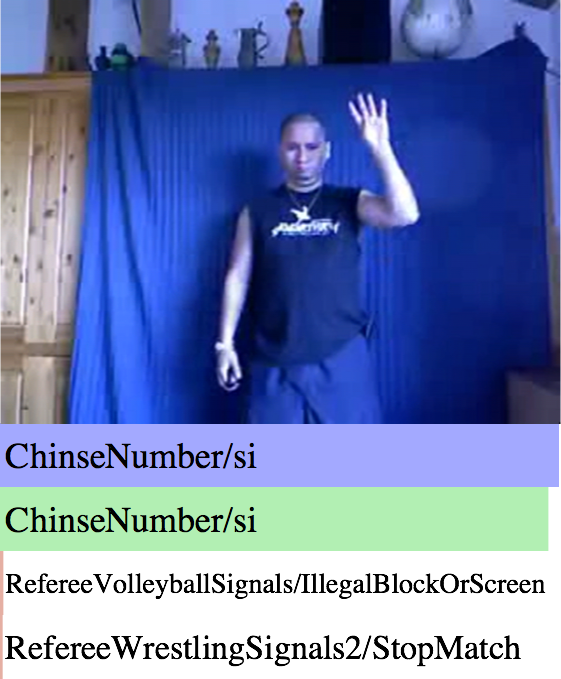}}
    \subfigure{
    \label{re2}
    \includegraphics[width=0.22\textwidth]{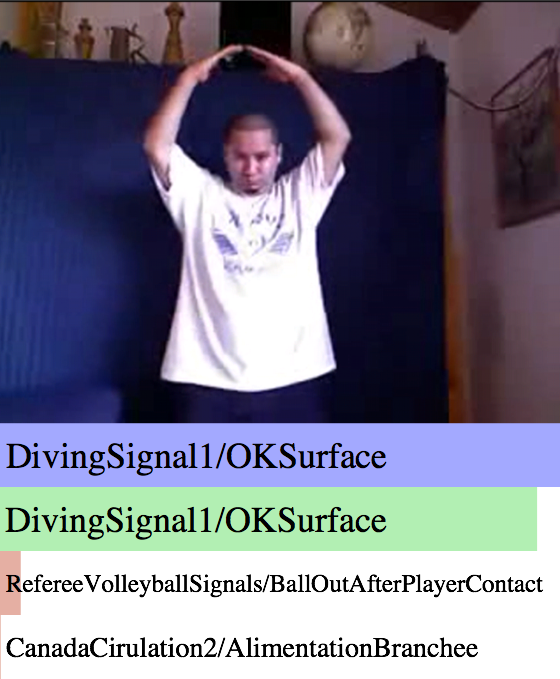}}
    \subfigure{
    \label{re3}
    \includegraphics[width=0.22\textwidth]{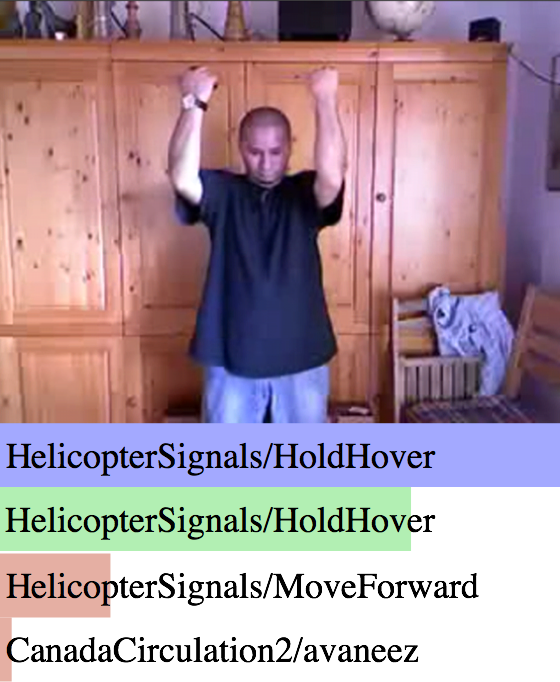}}
    \subfigure{
    \label{re1}
    \includegraphics[width=0.22\textwidth]{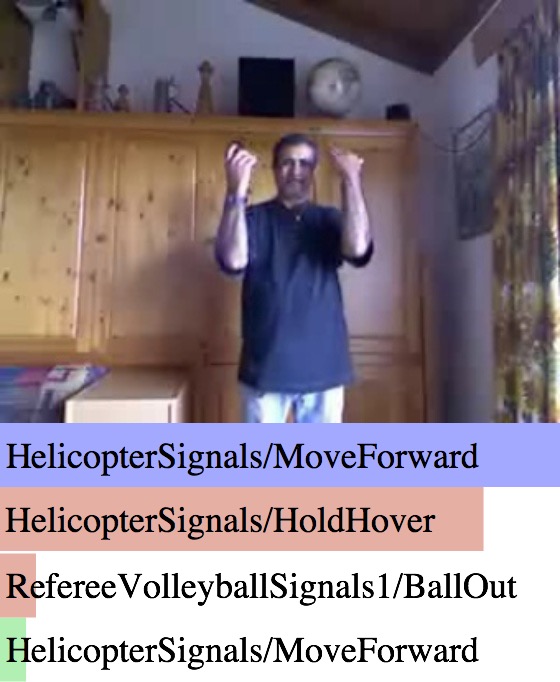}} \\


  \vspace{-2ex}
  \hspace{4ex}
    \subfigure{
    \label{re1}
    \includegraphics[width=0.22\textwidth]{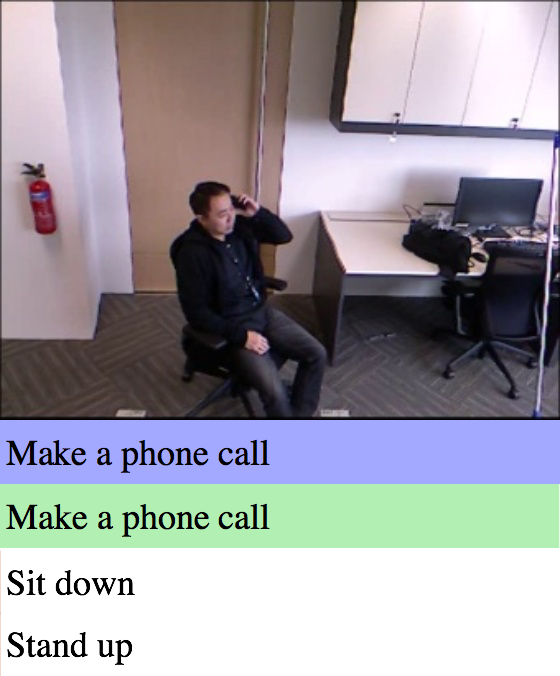}}
    \subfigure{
    \label{re2}
    \includegraphics[width=0.22\textwidth]{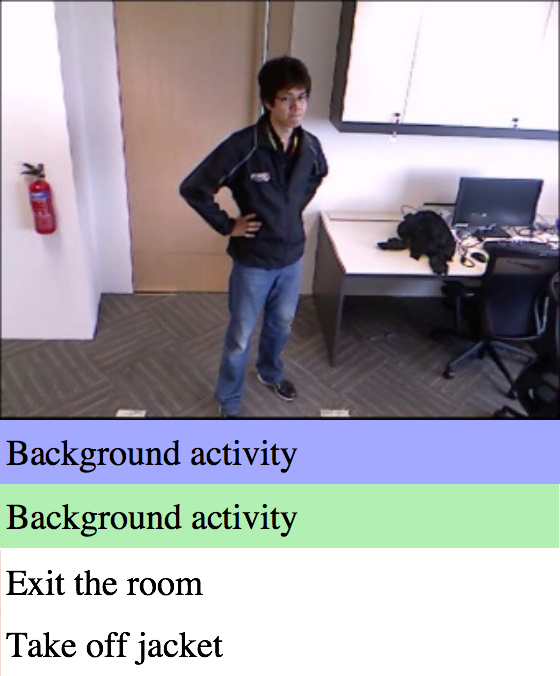}}
    \subfigure{
    \label{re3}
    \includegraphics[width=0.22\textwidth]{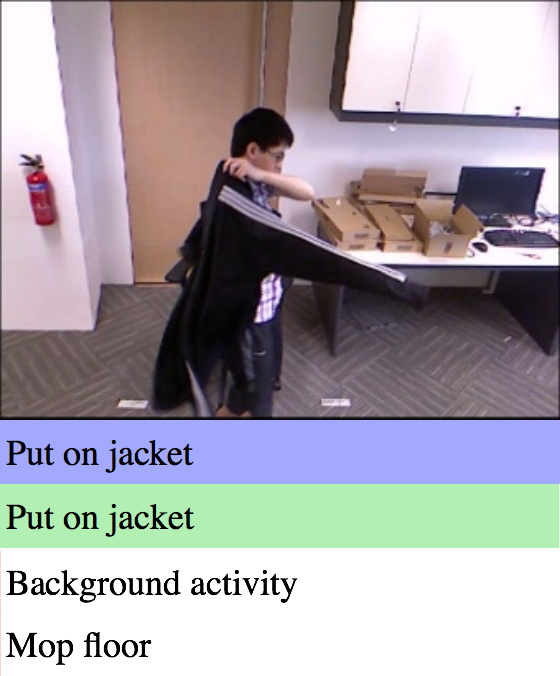}}
    \subfigure{
    \label{re1}
    \includegraphics[width=0.22\textwidth]{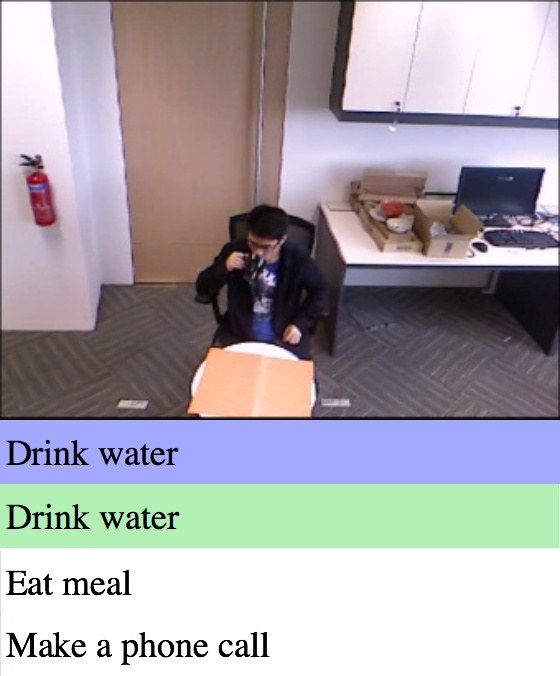}} \\
  \caption{Qualitative recognition results of our proposed approach on ChaLearn IsoGD (1st row) and RGB-D HuDaAct (2nd row) benchmarks. Bars colored blue indicate ground truths while green indicate correct and red wrong. The length of the bar represents confidence.}
  \label{fig:result}
\end{figure*}

Example recognition results are shown in Fig.\ref{fig:result} where the prediction distribution together with its confidence is displayed. We also show the ground-truth and top-3 predicted labels of each recognition result. As can be seen from the figure, our proposed approach correctly recognizes most of the gestures and attains pretty good accuracy even under challenging scenarios. However, the forth video in the first row is mis-classified because the first prediction (4th video) is very similar to ground-truth (3rd video).

\begin{figure}[!htb]
  \hspace{-2ex}
  \subfigure{
    \label{re1}
    \includegraphics[width=0.49\textwidth]{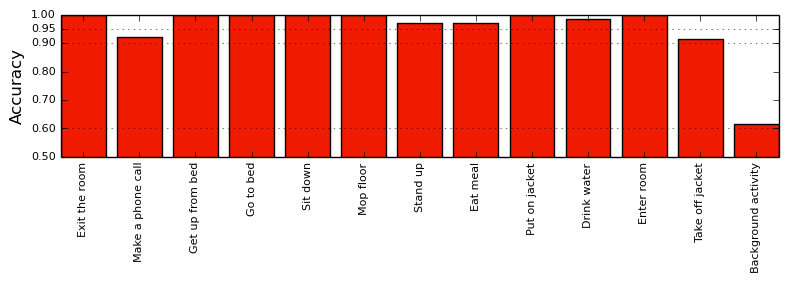}}
   \caption{Qualitative recognition results of our proposed approach on RGBD-HuDaAct benchmark}
  \label{fig:huda-acc}
\end{figure}

Fig.\ref{fig:huda-acc} displays the recognition result of each category in RGBD-HuDaAct, where the first nine classes achieve an average accuracy of over 90\%. For accuracies of different classes on ChaLearn IsoGD, please infer our supplementary material.

\subsection{Comparison with State of the Art}

\begin{table*}[!htb]\scriptsize
 \tabcolsep 10pt \caption{Comparison with state-of-the-art methods on ChaLearn IsoGD and RGBD-HuDaAct benchmarks(\%)}
  \begin{center}
  \begin{tabular}{| c | c | c | c || c  c c  c |}
    \hline
    \multicolumn{4}{|c||}{ChaLearn IsoGD Dataset} & \multicolumn{4}{c|}{RGBD-HuDaAct Dataset}\\
    \hline
     Method     & Result   & Method     & Result   & \multicolumn{3}{c|}{Method}    & Result   \\
    \hline
    NTUST           &20.33\%  &  MFSK+DeepID~\cite{jun-CVPRW}      & 23.67\%      &\multicolumn{3}{c|}{STIPs(K=512)~\cite{Laptev2003Space}}  & 79.77\%       \\
    MFSK~\cite{wan2016explore}            & 24.19\% &TARDIS             &40.15\%       &\multicolumn{3}{c|}{DLMC-STIPs(M =8)~\cite{hudaact}} & 79.49 \%   \\
    XJTUfx          & 43.92\% &ICT\_NHCI~\cite{ICT}          & 46.80\%      & \multicolumn{3}{c|}{DLMC-STIPs(K=512,SPM)~\cite{hudaact}} & 81.48\%    \\
     XDETVP-TRIMPS       &  50.93\%  &AMRL~\cite{wangpichao}               & 55.57\%      & \multicolumn{3}{c|}{3D-MHIs(Linear)~\cite{Davis2000The,hudaact}}       &70.51\%    \\
     FLiXT       & 56.90\%   & -                  &-             & \multicolumn{3}{c|}{3D-MHIs(RBF)~\cite{Davis2000The,hudaact}}          &69.66\%  \\
   \hline
   2SCVN-RGB (Ours)    & 45.65\%          &3DDSN-Sal (Ours)    &  43.35\%  & 2SCVN-RGB  &83.91\%   & 2SCVN-Flow  &  95.32\% \\
   2SCVN-Flow (Ours) & 58.36\%   &3DDSN-Depth (Ours)    & 54.95\%       & 3DDSN-Depth  &92.26\% &3DDSN-Sal   &92.06\% \\
   2SCVN-Fusion (Ours) &62.72\%  &3DDSN-Fusion(Ours)    &56.37\%       &2SCVN-Fusion &96.13\% &3DDSN-Fusion &93.68\% \\
   2SCVN-3DDSN (Ours) & \textbf{67.19\%}   & - &-        &2SCVN-3DDSN & \textbf{96.74\%}  & - &-     \\
   \hline
  \end{tabular}
  \end{center}
  \label{tab:all-result}
\end{table*}

We compare our proposed approach with competitors ranking top on the leaderboard of ChaLearn IsoGD benchmark~\cite{chalearn,jun-CVPRW} and state-of-the-art results on RGB-HuDaAct~\cite{hudaact} datasets. We also tested each modality of our proposed framework as well as their combinations. Final results are summarized in Table~\ref{tab:all-result}.

On ChaLearn IsoGD, hand-crafted features such as MFSK~\cite{wan2016explore} as well as its variant which combines DeepID feature~\cite{jun-CVPRW} scores relatively low compared to deep learning based methods such as AMRL~\cite{wangpichao} which incorporates three representations DDI, DDNI and DDMNI based on bidirectional rank pooling and ICT~\cite{ICT} which trains a two-stream RNN for RGB and depth stream respectively.

2SCVN-RGB achieves an accuracy of 45.65\% which is pretty good considering that it only uses one modalitity and that it only encodes static information. 2SCVN-Flow acquires a huge gain in performance as it captures motion information through stacked optical flow fields, which is reasonable because the accuracy of video recognition relies on the extent of understanding of the whole sequence. This is also what motivates us to explore consensus voting, a strategy that models the long term structure of the whole sequence to reduce estimation variance. The accuracy of 2SCVN-Fusion is further boosted as it combines the merits from spatial (2SCVN-RGB) and temporal stream (2SCVN-Flow). The performance of 3DDSN-Depth and 3DDSN-Sal are really impressive as they all score high compared to competing algorithms, due to rich representation capability of 3D convolution. Finally, although 2SCVN-Fusion scores rather high, the performance gain brought about 3DDSN after integration is still remarkable (over 6\%). This is in accordance with our observation that depth and saliency is supplementary to RGB modality. As can seen from Table~\ref{tab:all-result}, our proposed approach outperforms other competing algorithms by a large margin with over 10\% and 15\% accuracy on ChaLearn IsoGD and RGB-D HuDaAct respectively.

\section{Conclusions and Future Work}
\label{sec:conclusion}
In this paper, we have proposed a multi-modality framework for RGB-D gesture recognition that achieves superior recognition accuracies. Specifically, 2SCVN based on the strategy of consensus voting is employed to model long term video structure and reduce estimation variance while 3DDSN composed of depth and saliency streams are aggregated in parellel to capture embedded information supplementary to RGB modality. 3D-RGB stream is not adopted as it is inferior to 2SCVN. We also notice the possibility of employing 2SCVN on other modalities such as depth, depth-flow or saliency and we leave it as future work. Extensive experiments show the effectiveness of our framework and codes would be realsed to facilitate future research.

{\small
\bibliographystyle{ieee}
\bibliography{egpaper_final}
}

\end{document}